\def\year{2021}\relax
\documentclass[letterpaper,10pt]{article} % DO NOT CHANGE THIS
\usepackage{aaai21}  % DO NOT CHANGE THIS
\usepackage{times}  % DO NOT CHANGE THIS
\usepackage{helvet} % DO NOT CHANGE THIS
\usepackage{courier}  % DO NOT CHANGE THIS
\usepackage[hyphens]{url}  % DO NOT CHANGE THIS
\usepackage{graphicx} % DO NOT CHANGE THIS
\usepackage{natbib}  % DO NOT CHANGE THIS AND DO NOT ADD ANY OPTIONS TO IT
\usepackage{caption} % DO NOT CHANGE THIS AND DO NOT ADD ANY OPTIONS TO IT
\usepackage{threeparttable}
\usepackage{multirow}
\usepackage{booktabs} 
\usepackage{amssymb}
\usepackage{amsmath}
\usepackage{tabu}
\usepackage{times,mathptmx}
\usepackage{colortbl}
\usepackage{nameref}
\usepackage[textsize=tiny]{todonotes}
\newcommand{\ignore}[1]{}

\title{A Lightweight Neural Model for Biomedical Entity Linking}
\author {
    Lihu Chen,\textsuperscript{\rm 1}
    Gaël Varoquaux,\textsuperscript{\rm 2}
    Fabian M. Suchanek,\textsuperscript{\rm 1} \\
}
\affiliations {
    \textsuperscript{\rm 1} LTCI \& Télécom Paris \& Institut Polytechnique de Paris, France \\
    \textsuperscript{\rm 2} Inria \& CEA \& Université Paris-Saclay, France \\
    lihu.chen@telecom-paris.fr, suchanek@telecom-paris.fr, gael.varoquaux@inria.fr
}

\begin{document}
\maketitle
\begin{abstract}
Biomedical entity linking aims to map biomedical mentions, such as diseases and drugs, to standard entities in a given knowledge base. 
The specific challenge in this context is that the same biomedical entity can have a wide range of names,  including synonyms, morphological variations, and names with different word orderings. 
Recently, BERT-based methods have advanced the state-of-the-art by allowing for rich representations of word sequences.
However, they often have hundreds of millions of parameters and require heavy computing resources, which limits their applications in resource-limited scenarios.
Here, we propose a lightweight neural method for biomedical entity linking, which needs just a fraction of the parameters of a BERT model and much less computing resources. 
Our method uses a simple alignment layer with attention mechanisms to capture the variations between mention and entity names.
Yet, we show that our model is competitive with previous work on standard evaluation benchmarks.
\end{abstract}

\section*{Introduction}
% Fabian: Describing what it is
Entity linking (Entity Normalization) is the task of mapping entity mentions in text documents to standard entities in a given knowledge base. For example, the word ``Paris'' is \emph{ambiguous}: It can refer either to the capital of France or to a hero of Greek mythology. Now given the text ``Paris is the son of King Priam'', the goal is to determine that, in this sentence, the word refers to the Greek hero, and to link the word to the corresponding entity in a knowledge base such as YAGO \cite{suchanek2007yago} or DBpedia \cite{auer2007dbpedia}. 
%Intriguingly, the Greek hero also goes by the name of ``Alexander''. Thus, the words ``Paris'' and ``Alexander'' are \emph{synonymous}, and if they both refer to the Greek hero in some input text, they both have to be linked to the same entity in the knowledge base.

% Fabian: Describing why it's important
In the biomedical domain, entity linking maps mentions of diseases, drugs, and measures to normalized entities in standard vocabularies. It is an important ingredient for automation in medical practice, research, and public health.
Different names of the same entities in Hospital Information Systems seriously hinder the integration and use of medical data. If a medication appears with different names, researchers cannot study its impact, and patients may erroneously be prescribed the same medication twice. 

% Fabian: Describing why it's difficult
The particular challenge of biomedical entity linking is not the ambiguity: a word usually refers to only a single entity. Rather, the challenge is that the surface forms vary markedly, due to abbreviations, morphological variations, synonymous words, and different word orderings. 
For example, \textit{``Diabetes Mellitus, Type 2''} is also written as \textit{``DM2''} and \textit{``lung cancer''} is also known as \textit{``lung neoplasm malignant''}. In fact, the surface forms vary so much that all the possible expressions of an entity cannot be known upfront. This means that standard disambiguation systems cannot be applied in our scenario, because they assume that all forms of an entity are known.
%, and thus they cannot be applied in our scenario.

One may think that variation in surface forms is not such a big problem, as long as all variations  of an entity are sufficiently close to its canonical form. Yet, this is not the case.
For example, the phrase \textit{"decreases in hemoglobin"} could refer to at least 4 different entities in MedDRA, which all look alike:  \textit{"changes in hemoglobin"}, \textit{"increase in hematocrit"}, \textit{"haemoglobin decreased"}, and \textit{"decreases in platelets"}.
In addition, biomedical entity linking cannot rely on external resources such as 
alias tables, entity descriptions, or entity co-occurrence, which are often used in classical entity linking settings.

% Fabian: what has been done
For this reason, entity linking approaches have been developed particularly for biomedical entity linking.
Many methods use deep learning: the work of \citet{li2017cnn} casts biomedical entity linking as a ranking problem,  leveraging convolutional neural networks (CNNs). 
More recently, the introduction of BERT has advanced the performance of many NLP tasks, including in the biomedical domain \cite{huang2019clinicalbert,lee2020biobert,ji2020bert}. 
BERT creates rich pre-trained representations on unlabeled data and achieves
state-of-the-art performance on a large suite of sentence-level and token-level tasks, outperforming many task-specific architectures.
However, considering the number of parameters of pre-trained BERT models, 
the improvements brought by fine-tuning them come with a heavy computational cost and memory footprint. 
This is a problem for energy efficiency, for smaller organizations, or in poorer countries.

In this paper, we introduce a very lightweight model that achieves a performance statistically indistinguishable from the state-of-the-art BERT-based models. The central idea is to use an alignment layer with an attention mechanism, 
which can capture the similarity and difference of corresponding parts between candidate and mention names. Our model is 23x smaller and 6.4x faster than BERT-based models on average; and more than twice smaller and faster than the lightweight BERT models. Yet, as we show, our model achieves comparable performance on all standard benchmarks. Further, we can show that adding more complexity to our model is not necessary: the entity-mention priors, the context around the mention, or the coherence of extracted entities \cite[as used, e.g., in][]{hoffart2011robust} do not improve the results any further.
\footnote{All data and code are available at  \url{https://github.com/tigerchen52/Biomedical-Entity-Linking}.}

\section*{Related Work}

In the biomedical domain, much early research focuses on capturing string similarity of mentions and entity names with rule-based systems~\cite{dogan2012inference, kang2013using, d2015sieve}.
Rule-based systems are simple and transparent, but researchers need to define rules manually, and these are bound to an application.

To avoid manual rules, machine-learning approaches learn suitable similarity measures between mentions and entity names automatically from training sets~\cite{leaman2013dnorm, dougan2014ncbi, ghiasvand2014r, leaman2016taggerone}. 
However, one drawback of these methods is that they cannot recognize semantically related words.
%For example, they cannot see that `cardiac' and `heart' are closely related.

Recently, deep learning methods have been successfully applied to different NLP tasks, based on 
pre-trained word embeddings, such as word2vec \cite{mikolov2013distributed} and Glove \cite{pennington2014glove}. 
\citet{li2017cnn} and \citet{wright2019normco} introduce a CNN and RNN, respectively, with pre-trained word embeddings, which casts biomedical entity linking into a ranking problem.

However, traditional methods for learning word embeddings allow for only a single context-independent representation of each word.  
Bidirectional Encoder Representations from Transformers (BERT) address this problem by pre-training deep bidirectional representations from unlabeled text, jointly conditioning on both the left and the right context in all layers.
\citet{ji2020bert} proposed an biomedical entity normalization architecture by fine-tuning the pre-trained BERT / BioBERT / ClinicalBERT models \cite{devlin2018bert,huang2019clinicalbert,lee2020biobert}.  
Extensive experiments show that their model outperforms previous methods and advanced the state-of-the-art for biomedical entity linking. A shortcoming of BERT is that it needs high-performance machines.

\section*{Our Approach}
Formally, our inputs are (1) a \emph{knowledge base} (KB), i.e., a list of entities, each with one or more names, and (2) a \emph{corpus}, i.e., a set of text documents in which certain text spans have been tagged as entity mentions. The goal is to link each entity mention to the correct entity in the KB. To solve this problem, we are given a training set, i.e., a part of the corpus where the entity mentions have been linked already to the correct entities in the KB.
Our method proceeds in 3 steps:

\begin{description}
	\item[\textbf{Preprocessing.}] We preprocess all mentions in the corpus and entity names in the KB to bring them to a uniform format.
	\item[\textbf{Candidate Generation.}] For each mention, we generate a set of candidate entities from the KB.
	\item[\textbf{Ranking Model.}] For each mention with its candidate entities, we use a ranking model to score each pair of mention and candidate, outputting the top-ranked result.
\end{description}

\noindent Let us now describe these steps in detail.

\subsection*{Preprocessing}
We preprocess all mentions in the corpus and all entity names in the KB by the following steps:

\textbf{Abbreviation Expansion.} Like previous work~\cite{ji2020bert}, we use the Ab3p Toolkit~\cite{sohn2008abbreviation} to expand medical abbreviations. The Ab3p tool outputs a probability for each possible expansion, and we use the most probable expansion. For example, Ab3p knows that ``DM'' is an abbreviation of ``Diabetes Mellitus'', and so we replace the abbreviation with its expanded term.
We also expand mentions by the first matching one from an abbreviation dictionary constructed by previous work \cite{d2015sieve}, and supplement 20 biomedical abbreviations manually (such as Glycated hemoglobin (HbA1c)). Our dictionary is available in the supplementary material and online.

\textbf{Numeral Replacement.} Entity names may contain numerals in different forms (e.g., Arabic, Roman, spelt out in English, etc.)
We replace all forms with spelled-out English numerals. For example, ``type \uppercase\expandafter{\romannumeral2} diabetes  mellitus'' becomes ``type two diabetes  mellitus''. For this purpose, we manually compiled a dictionary of numerals from the corresponding Wikipedia pages. Finally, we remove all punctuation, and convert all words to lowercase.
\begin{figure*}[t]
	\centering
	\includegraphics[width=0.8\textwidth]{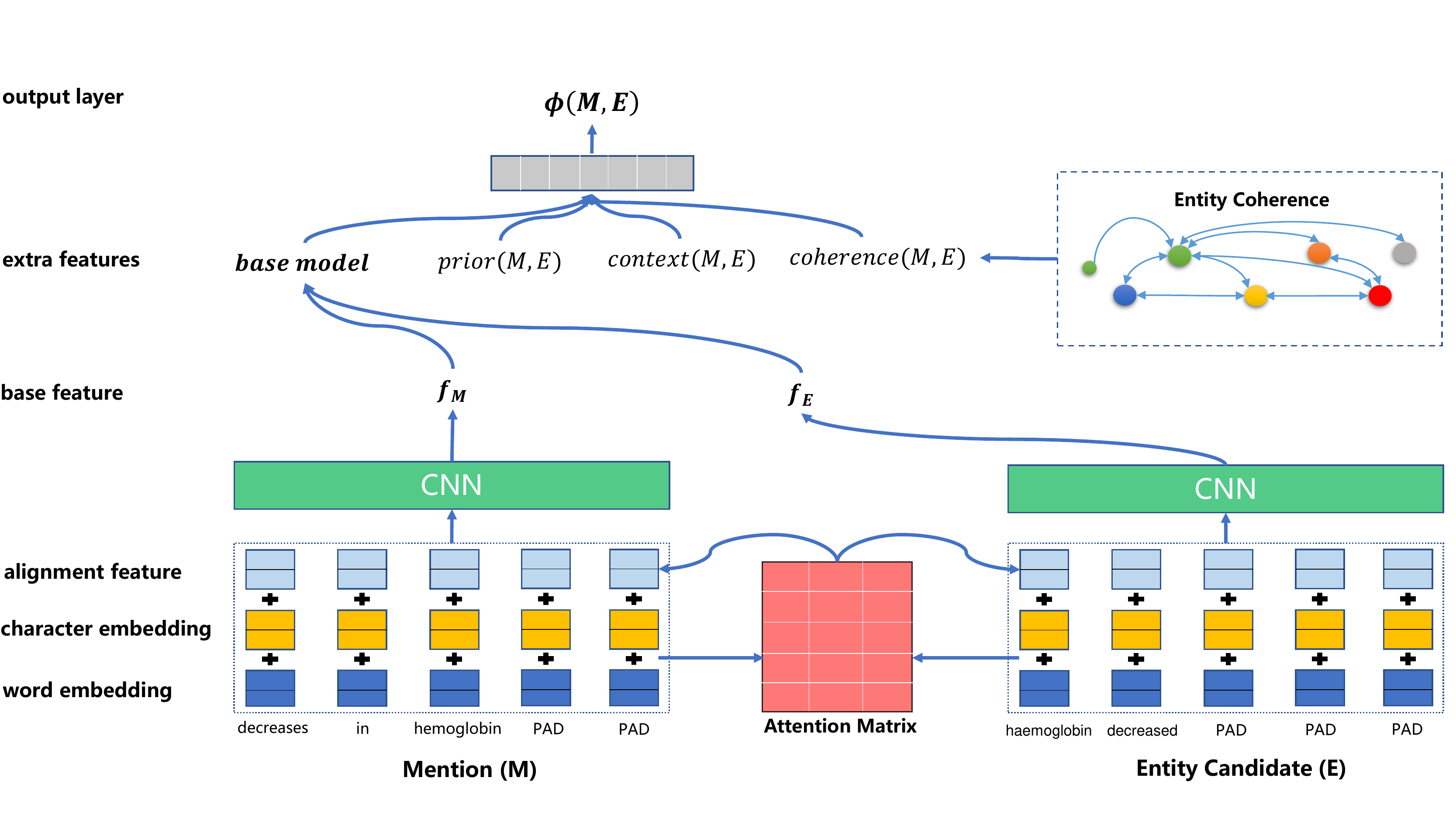}
	\caption{The architecture of our ranking model, with the input mention ``decreases in hemoglobin'' and the input entity candidate ``haemoglobin  decreased''.}
	\label{fig:architecture}
\end{figure*}

\textbf{KB Augmentation.} We augment the KB by adding all names from the training set to the corresponding entities. For example, if the training set links the mention ``GS'' in the corpus to the entity ``Adenomatous polyposis coli'' in the KB, we add ``GS'' to the names of that entity in the KB. 

\subsection*{Candidate Generation}\label{sec:cand}
Our ranking approach is based on a deep learning architecture that can compute a similarity score for each pair of a mention in the corpus and an entity name in the KB. However, it is too slow to apply this model to all combinations of all mentions and all entities. 
Therefore, we generate, for each mention $M$ in the corpus, a set $C_M$ of candidate entities from the KB. Then we apply the deep learning method only to the set $C_M$.

To generate the candidate set $C_M$, we calculate a score for $M$ and each entity in the KB, and return the top-$k$ entities with the highest score as the candidate set $C_M$ (in our experiments, $k=20$).
As each entity has several names, we calculate the score of $M$ and all names of the entity $E$, and use the maximum score as the score of $M$ and the entity $E$.

To compute the score between a mention $M$ and an entity name $S$, we split each of them into tokens, so that we have $M=\{m_{1}, m_{2},..., m_{|M|}\}$ 
and $S=\{s_{1}, s_{2},..., s_{|S|}\}$.

We represent each token by a vector taken from pre-trained embedding matrix $\mathbf V \in \mathbb{R}^{d\times | V  |}$ where $d$ is the dimension of word vectors and $V$ is a fixed-sized vocabulary (details in the section of \nameref{sec:experimental setting}).
To take into account the possibility of different token orderings in $M$ and $S$, we design the \emph{aligned cosine similarity} (\textit{ACos}), which maps a given token $m_i \in M$ to the most similar token $s_j \in S$ and returns the cosine similarity to that token:
\begin{equation}
\textit{ACos}(m_{i}, S) = \max \{ cos(m_{i}, s_{j}) \mid s_{j} \in S \}   
\end{equation}

\noindent The similarity score is then computed as the sum of the aligned cosine similarities. To avoid tending to long text, and to make the metric symmetric, we add the similarity scores in the other direction as well, yielding:

\begin{multline}
\textit{sim}(M,S) = 
\frac{1}{\left| M \right| + \left| S \right|} (\sum_{m_{i} \in M} \textit{ACos}(m_{i}, S) \\
 + \sum_{s_{j} \in S} \textit{ACos}(s_{j},M))
\end{multline} 

\noindent We can now construct the candidate set $C_M = \{\langle{}E_{1}, S_{1}\rangle,$ $\langle{}E_{2}, S_{2}\rangle,$ $..., \langle{}E_{k}, S_{k}\rangle\}$ where $E_i$ is the id of the entity, and $S_i$ is the chosen name of the entity. 
This set contains the top-$k$ ranked entity candidates for each mention $M$.
Specifically, if there are candidates whose score is equal to 1 in this set, 
we will filter out other candidates whose score is less than 1.

\subsection*{Ranking Model}
Given a mention $M$ and its candidate set $C_M = \{\langle{}E_{1}, S_{1}\rangle,$ $\langle{}E_{2}, S_{2}\rangle,$ $..., \langle{}E_{k}, S_{k}\rangle\}$, the ranking model computes a score for each pair of the mention and an entity name candidate $S_i$. %: $\{ <M, E_{1}>, <M, E_{2}>,..., <M, E_{n}>\}$. 
Figure~\ref{fig:architecture} shows the corresponding neural network architecture. Let us first describe the base model.
This model relies exclusively on the text similarity of mentions and entity names. It ignores the context in which a mention appears, or the prior probability of the target entities. To compute the text similarity, we crafted the neural network following the candidate generation: it determines, for each token in the mention, the most similar token in the entity name, and vice versa. Different from the candidate generation, we also take into account character level information here and use an alignment layer to capture the similarity and difference of correspondences between mention and entity names.

\paragraph{Representation Layer.}
As mentioned in the \nameref{sec:cand}, we represent a mention $M$ and an entity name $S$ by the set of the embeddings of its tokens in the vocabulary $V$.
However, not all tokens exist in the vocabulary $V$.
To handle out-of-vocabulary words, we adopt a recurrent Neural Network (RNN) to capture character-level features for each word.
This has the additional advantage of learning the morphological variations of words.
We use a Bi-directional LSTM (BiLSTM), running a forward and backward LSTM on a character sequence \cite{graves2013speech}.  
We concatenate the last output states of these two LSTMs as the character-level representation of a word.
To use both word-level and character-level information, we represent each token of a mention or entity name as the concatenation of its embedding in $V$ and its character-level representation.

\paragraph{Alignment Layer.} To counter the problem of different word orderings in the mention and the entity name, we want the network to find, for each token in the mention, the most similar token in the entity name.
For this purpose, we adapt the attention mechanisms that have been developed for
machine comprehension and answer selection~\cite{chen2016enhanced,wang2016compare}. 

Assume that we have a mention $M = \{\bar{m}_{1},$ $\bar{m}_{2},$ $..., \bar{m}_{|M|}\}$ and an entity name $S = \{\bar{s}_{1},$ $\bar{s}_{2},$ $..., \bar{s}_{|S|}\}$, which were generated by the Representation Layer. 
We calculate a $|M|\times|S|$-dimensional weight matrix $W$, whose element $w_{i,j}$ indicates the similarity between the token $i$ of the mention and the token $j$ of the entity name, $w_{ij} = \bar{m}_{i}^{T} \bar{s}_{j}$. Thus, the $i^{th}$ row in $W$ represents the similarity between the $i^{th}$ token in $M$ and each token in $S$. We apply a softmax function on each row of $W$ to normalize the values, yielding a matrix $W'$. We can then compute a vector $\tilde{m}_i$ for the $i^{th}$ token of the mention, which is the sum of the vectors of the tokens of $S$, weighted by their similarity to $\bar{m}_i$:

\begin{equation}
    \tilde{m}_{i} = \sum_{j=1}^{t} w_{ij}' \bar{s}_{j}
\end{equation}

\noindent This vector ``reconstructs'' $\bar{m}_i$ by adding up suitable vectors from $S$, using mainly those vectors of $S$ that are similar to $\bar{m}_i$. If this reconstruction succeeds (i.e., if $\bar{m}_i$ is similar to $\tilde{m}_i$), then $S$ contained tokens which, together, contain the same information as $\bar{m}_i$. 
\ignore{
it so that we obtain an attention matrix where each element $\alpha_{ij} \in [0, 1]$: 
\begin{equation}
\alpha_{ij} = \frac{exp ( w_{ij}  )}{\sum_{k=1}^{t} w_{ik}}
\end{equation}
while we also apply a softmax function on each column of $W$ to get the attention matrix for $S$:
\begin{equation}
\beta_{ij} = \frac{exp ( w_{ij}  )}{\sum_{k=1}^{t} w_{lj}}
\end{equation}
After, the alignment representation can be computed as a weighted sum:
\begin{align}
\tilde{m}_{i} = \sum_{j=1}^{t}\beta_{ij} \bar{s}_{j}
&&\text{and}&&
\tilde{s}_{j} = \sum_{i=1}^{l}\alpha_{ij} \bar{m}_{i}
\end{align}
where $\tilde{m}_{i}$ is the most relevant part to $\bar{m}_{i}$ that selected from $ S = \{ \bar{s}_{1}, \bar{s}_{2},..., \bar{s}_{t}\}$.
We do the same operation for each word in $S$ to get $\tilde{s}_{j}$. 
In this step, we can find the corresponding parts of two texts to compare without being influenced by the order of words
}
To measure this similarity, we could use a simple dot-product. However, this reduces the similarity to a single scalar value, which erases precious element-wise similarities. Therefore, we use the following two comparison functions
%, which compute in an element-wise level. These comparison function has been explored by previous work
\cite{tai2015improved,wang2016compare}:
\begin{equation}
    \textit{sub}(\bar{m}_{i}, \tilde{m}_{i}) = (\bar{m}_{i}-\tilde{m}_{i}) \odot (\bar{m}_{i}-\tilde{m}_{i})
\end{equation}

\begin{equation}
    \textit{mul}(\bar{m}_{i}, \tilde{m}_{i}) =  \bar{m}_{i} \odot \tilde{m}_{i}
\end{equation}

\noindent where the operator $\odot$ means element-wise multiplication. Intuitively, the functions $sub$ and $mul$ represent subtraction and multiplication, respectively. 
The function \emph{sub} has similarities to the Euclidean distance, while \emph{mul} has similarities to the cosine similarity -- while preserving the element-wise information.
Finally, we obtain a new representation of each token $i$ of the mention by concatenating $\bar{m}_{i}, \tilde{m}_{i}$ and their difference and similarity:
\begin{equation}
    \hat{m}_{i} = [\bar{m}_{i}, \tilde{m}_{i}, \textit{sub}(\bar{m}_{i}, \tilde{m}_{i}), \textit{mul}(\bar{m}_{i}, \tilde{m}_{i}) ]
\end{equation}

\noindent By applying the same procedure on the columns of $W$, we can compute analogously a vector $\tilde{s}_{j}$ for each token vector $s_j$ of $S$, and obtain the new representation for the $j^{th}$ token of the entity name as
\begin{equation}
    \hat{s}_{j} = [\bar{s}_{j}, \tilde{s}_{j}, \textit{sub}(\bar{s}_{j}, \tilde{s}_{j}), \textit{mul}(\bar{s}_{j}, \tilde{s}_{j}) ]
\end{equation}

\noindent This representation augments the original representation $\bar{s}_{j}$ of the token by the ``reconstructed'' token $\tilde{s}_{j}$, and by information about how similar $\tilde{s}_{j}$ is to $\bar{s}_{j}$.

\paragraph{CNN Layer.}
We now have rich representations for the mention and the entity name, and we apply a one-layer CNN on the mention $[\hat{m}_{1}, \hat{m}_{2},..., \hat{m}_{|M|}]$ and the entity name $[\hat{s}_{1}, \hat{s}_{2},..., \hat{s}_{|S|}]$. We adopt the CNN architecture proposed by \cite{kim2014convolutional} to extract n-gram features of each text:
\begin{equation}
    f_{M} = \textit{CNN}([\hat{m}_{1}, \hat{m}_{2},..., \hat{m}_{M}])
\end{equation}

\begin{equation}
   f_{E} = \textit{CNN}([\hat{s}_{1}, \hat{s}_{2},..., \hat{s}_{S}])
\end{equation}

\noindent We concatenate these to a single vector $f_{\textit{out}} = [ f_{M}, f_{E}  ]$.

\paragraph{Output Layer.}
We are now ready to compute the final output of our network
using a two-layer fully connected neural network:
\begin{equation}
    \Phi  ( M, E  ) = \textit{sigmoid} (W_{2}~~\textit{ReLU}(W_{1}~f_{\textit{out}} + b_{1} ) + b_{2} )
\end{equation}

\noindent where $W_{2}$ and $W_{1}$ are learned weight matrices, and $b_1$ and $b_2$ are bias values.
This constitutes our base model, which relies solely on string similarity. 
We will now see how we can add add prior, context, and coherence features. 

\subsection*{Extra Features}\label{sec:extra}

\paragraph{Mention-Entity Prior.}
Consider an ambiguous case such as \textit{``You should shower, let water flow over wounds, pat dry with a towel.''} appearing in hospital Discharge Instructions. In this context, the disease name \textit{``wounds''} is much more likely to refer to \textit{``surgical wound''} than \textit{``gunshot wound''}. This prior probability is called the \emph{mention-entity prior}. It can be estimated, e.g., 
by counting in Wikipedia how often a mention is linked to the page of an entity~\cite{hoffart2011robust}.
Unlike DBpedia and YAGO, biomedical knowledge bases generally do not provide links to Wikipedia.  
Hence, we estimate the mention-entity prior from the training set, as: 
\begin{equation}
   \textit{prior}(M,E) = \log \textit{count}(M, E)
\end{equation}

\noindent where $\textit{count}(M, E)$ is the frequency with which the mention $M$ is linked to the target entity $E$ in the training dataset. To reduce the effect of overly large values, we apply the logarithm.
This prior can be added easily to our model by concatenating it in $f_{\textit{out}}$: 
\begin{equation}
f_{\textit{out}} = [ f_{M}, f_{E}, \textit{prior}(M,E) ]
\end{equation}

\paragraph{Context.}
The context around a mention can provide clues on which candidate entity to choose. 
We compute a context score that measures how relevant the keywords of the context are to the candidate entity name.    
We first represent the sentence containing the mention by pre-trained word embeddings. We then run a Bi-directional LSTM on the sentence to get a new representation for each word.
In the same way, we apply a Bi-directional LSTM on the entity name tokens to get the entity name representation $cxt_{E}$.
To select keywords relevant to the entity while ignoring noise words, we adopt an attention strategy to assign a weight for each token in the sentence.
Then we use a weighted sum to represent the sentence as  $cxt_{M}$.
The context score is then computed as the cosine similarity between both representations:
\begin{equation}
\textit{context}(M, E) = \cos (cxt_{M}, cxt_{E})
\end{equation}
As before, we concatenate this score to the vector $f_{out}$.

\paragraph{Coherence.}
Certain entities are more likely to occur together in the same document than others, and we can leverage this disposition to help the entity linking.
To capture the co-occurrence of entities, we pre-train entity embeddings in such a way that entities that often co-occur together have a similar distributed representation.
We train these embeddings with Word2Vec~\cite{mikolov2013distributed} on a collection of PubMed abstracts\footnote{ftp://ftp.ncbi.nlm.nih.gov/pubmed/baseline/}. 
Since the entities in this corpus are not linked to our KB, we consider every occurrence of an exact entity name as a mention of that entity.

Given a mention $M$ and a candidate entity $E$, we compute a coherence score to measure how often the candidate entity co-occurs with the other entities in the document. We first select the mentions around $M$. 
For each mention, we use the first entity candidate (as given by the candidate selection).
This gives us a set of entities $P_{M} = \{ {p}_{1}, {p}_{2},..., {p}_{k}\}$, where each element is a pre-trained entity vector. Finally, the coherence score is computed as:  
\begin{equation}
\textit{coherence}(M, E) = \frac{1}{k} \sum_{i=1}^{k} \cos(p_{i},p_{E})
\end{equation}
\noindent where $p_{E}$ is the pre-trained vector of the entity candidate $E$. 
This score measures how close the candidate entity $E$ is, on average, to the other presumed entities in the document.
As before, we concatenate this score to the vector $f_{\textit{out}}$.
More precisely, we pre-trained separate entity embeddings for the three datasets and used the mean value of all entity embeddings to represent missing entities.

\subsection*{NIL Problem}
The NIL problem occurs when a mention does not correspond to any entity in the KB.
We adopt a traditional threshold method, which considers a mention unlinkable if its score is less than a threshold $\tau$.
This means that we map a mention to the highest-scoring entity if that score exceeds $\tau$, and to NIL otherwise. 
The threshold $\tau$ is learned from a training set.
For datasets that do not contain unlinkable mentions, we set the threshold $\tau$ to zero. 

\subsection*{Training}
For training, we adopt a triplet ranking loss function to make the score of the positive candidates higher than the score of the negative candidates.
The objective function is:
\begin{multline}
\theta ^{*} =  \mathop{\arg\min}_{\theta} \sum_{D \in \mathcal{D}}\sum_{M \in D}\sum_{E \in C} \\
 \max (0, \gamma + \Phi  ( M, E^{+}  ) - \Phi  ( M, E^{-}  ))
\end{multline}  
\noindent where $\theta$ stands for the parameters of our model.
$\mathcal{D}$ is a training set containing a certain number of documents and $\gamma$ is the parameter of margin. 
$E^{+}$ and $E^{-}$ represent a positive entity candidate and a negative entity candidate, respectively.
Our goal is to find an optimal $\theta$, which makes the score difference between positive and negative entity candidates as large as possible.
For this, we need triplets of a mention $M$, a positive example $E^+$ and a negative example $E^-$. The positive example can be obtained from the training set. 
The negative examples are usually chosen by random sampling from the KB.
In our case, we sample the negative example from the candidates that were produced by the candidate generation phase (excluding the correct entity).
This choice makes the negative examples very similar to the positive example, and forces the process to learn what distinguishes the positive candidate from the others.

\section*{Experiments}
\begin{table}[b!]
    %\fontsize{6.5}{8}\selectfont 
	\small
		\begin{tabu} {p{1.3cm} X[c] X[c] X[c] X[c] X[c] X[c]}
			\toprule
			&\multicolumn{2}{c}{ShARe/CLEF} &\multicolumn{2}{c}{NCBI} &\multicolumn{2}{c}{ADR} \\
			&train &test  &train  &test  &train  &test  \\
			\midrule
			documents &199 &99 &692 &100 &101 &99 \\
			mentions  &5816 &5351 &5921 &964 &7038 &6343 \\
			NIL    &1641 &1750 &0 &0 &47 &18 \\
			\midrule
			concepts &\multicolumn{2}{c}{88140} &\multicolumn{2}{c}{9656} &\multicolumn{2}{c}{23668}  \\
			synonyms &\multicolumn{2}{c}{42929} &\multicolumn{2}{c}{59280} &\multicolumn{2}{c}{0}\\
			\bottomrule
		\end{tabu}
	\caption{Dataset Statistics}\label{tab:datasets}
\end{table}

\ignore{
\begin{table*}[!t]  
	\small
	\begin{minipage}{.25\linewidth}
		\caption{Performance of different models. Results in gray are not statistically different from the top result.}\label{tab:performance_comparison}
	\end{minipage}%
	\hfill%
	\begin{minipage}{.72\linewidth}%
	\begin{threeparttable}
\begin{tabular}{cccccc}  
			\toprule  
			\multirow{1}{*}{Model}&ShARe/CLEF&NCBI&ADR\cr    
			\midrule
			DNorm \cite{leaman2013dnorm}&-&82.20$\pm$4.05&-\cr
			UWM \cite{ghiasvand2014r} &89.50$\pm$1.38&-&-\cr
			Sieve-based Model \cite{d2015sieve}&\cellcolor{lightgray!50}90.75$\pm$1.31&84.65$\pm$3.84&-\cr
			TaggerOne \cite{leaman2016taggerone}&-&\cellcolor{lightgray!50}88.80$\pm$3.32&-\cr
			Learning to Rank \cite{xu2017uth_ccb}&-&-&92.05$\pm$1.12\cr
			CNN-based Ranking \cite{li2017cnn}&\cellcolor{lightgray!50}90.30$\pm$1.33&86.10$\pm$3.63&-\cr
			BERT-based Ranking \cite{ji2020bert}&\cellcolor{lightgray!50}{\bf91.06$\pm$\bf1.29}&\cellcolor{lightgray!50}89.06$\pm$3.32&\cellcolor{lightgray!50}{\bf93.22$\pm$\bf1.04}\cr
			Our Base Model &\cellcolor{lightgray!50}90.10$\pm$1.35 &\cellcolor{lightgray!50}89.07$\pm$3.32&\cellcolor{lightgray!50}92.89$\pm$1.06\cr
			Our Base Model + Extra Features &\cellcolor{lightgray!50}90.43$\pm$1.33 &\cellcolor{lightgray!50}{\bf89.59$\pm$3.22}&\cellcolor{lightgray!50}93.00$\pm$1.06\cr
			\bottomrule  
		\end{tabular}
	\end{threeparttable}
	%\caption{Performance of different models. Results in gray are not statistically different from the top result.}\label{tab:performance_comparison}
\end{minipage}%
\end{table*}
}
\begin{table*}[!t]  
	
	\centering  
	%\fontsize{6.5}{8}\selectfont
	\small
	\begin{threeparttable}
		\begin{tabular}{cccccc}  
			\toprule  
			\multirow{1}{*}{Model}&ShARe/CLEF&NCBI&ADR\cr    
			\midrule
			DNorm \cite{leaman2013dnorm}&-&82.20$\pm$3.09&-\cr
			UWM \cite{ghiasvand2014r} &89.50$\pm$1.02&-&-\cr
			Sieve-based Model \cite{d2015sieve}&\cellcolor{lightgray!50}90.75$\pm$0.96&84.65$\pm$3.00&-\cr
			TaggerOne \cite{leaman2016taggerone}&-&\cellcolor{lightgray!50}88.80$\pm$2.59&-\cr
			Learning to Rank \cite{xu2017uth_ccb}&-&-&92.05$\pm$0.84\cr
			CNN-based Ranking \cite{li2017cnn}&\cellcolor{lightgray!50}90.30$\pm$1.00&86.10$\pm$2.79&-\cr
			BERT-based Ranking \cite{ji2020bert}&\cellcolor{lightgray!50}{\bf91.06$\pm$\bf0.96}&\cellcolor{lightgray!50}89.06$\pm$2.63&\cellcolor{lightgray!50}{\bf93.22$\pm$\bf0.79}\cr
			Our Base Model &\cellcolor{lightgray!50}90.10$\pm$1.00 &\cellcolor{lightgray!50}89.07$\pm$2.63&\cellcolor{lightgray!50}92.63$\pm$0.81\cr
			Our Base Model + Extra Features &\cellcolor{lightgray!50}90.43$\pm$0.99 &\cellcolor{lightgray!50}{\bf89.59$\pm$2.59}&\cellcolor{lightgray!50}92.74$\pm$0.80\cr
			\bottomrule  
		\end{tabular}
		
	\end{threeparttable}  
	\caption{Performance of different models. Results in gray are not statistically different from the top result.}\label{tab:performance_comparison}
\end{table*}

\subsection*{Datasets and Metrics. } 
We evaluate our model on three datasets (shown in Table~\ref{tab:datasets}). The \textbf{ShARe/CLEF} corpus~\cite{pradhan2013task} comprises 199 medical reports for training and 99 for testing. 
As Table~\ref{tab:datasets} shows, $28.2\%$ of the mentions in the training set and $32.7\%$ of the mentions in the test set are unlinkable.
The reference knowledge base used here is the SNOMED-CT subset of the UMLS 2012AA~\cite{bodenreider2004unified}.
The \textbf{NCBI} disease corpus~\cite{dougan2014ncbi} is a collection of 793 PubMed abstracts partitioned into 693 abstracts for training and development and 100 abstracts for testing.
We use the July 6, 2012 version of MEDIC~\cite{davis2012medic}, which contains 9,664 disease concepts.
The TAC 2017 Adverse Reaction Extraction (\textbf{ADR}) dataset consists of a training set of 101 labels and a test set of 99 labels.
The mentions have been mapped manually to the MedDRA 18.1 KB, which contains 23,668 unique concepts.

Following previous work, we adopt accuracy to compare the performance of different models.

\subsection*{Experimental Settings} \label{sec:experimental setting}
We implemented our model using Keras, and trained our model on a single Intel(R) Xeon(R) Gold 6154 CPU @ 3.00GHz, using less than 10Gb of memory. 
Each token is represented by a 200-dimensional word embedding computed on the PubMed and MIMIC-III corpora~\cite{zhang2019biowordvec}.
As for the character embeddings, we use a random matrix initialized as proposed in \citet{he2015delving}, with a dimension of $128$.
The dimension of the character LSTM is $64$, which yields $128$-dimensional character feature vectors.
In the CNN layer, the number of feature maps is $32$, and the filter windows are $[1, 2, 3]$. 
The dimension of the context LSTM and entity embedding is set to $32$ and $50$ respectively. 
We adopt a grid search on a hold-out set from training samples to select the value $\tau$, and and find an optimal for $\tau = 0.75$.

During the training phase, we select at most $20$ entity candidates per mention, and the parameter of the triplet rank loss is $0.1$.
For the optimization, we use Adam with a learning rate of $0.0005$ and a batch size of $64$.
To avoid overfitting, we adopt a dropout strategy with a dropout rate of $0.1$.

\subsection*{Competitors}
We compare our model to the following competitors:
\textbf{DNorm} \cite{leaman2013dnorm};
\textbf{UWM} \cite{ghiasvand2014r};
\textbf{Sieve-based Model} \cite{d2015sieve};
\textbf{TaggerOne} \cite{leaman2016taggerone};
a model based on \textbf{Learning to Rank} \cite{xu2017uth_ccb};
\textbf{CNN-based Ranking}  \cite{li2017cnn};
and \textbf{BERT-based Ranking} \cite{ji2020bert}.

\section*{Results} 

\subsection*{Overall Performance}
During the candidate generation, we generate 20 candidates for each mention. The recall of correct entities on the ShARe/CLEF, NCBI, and ADR test datasets is 97.79\%, 94.27\%, and 96.66\% respectively. We thus conclude that our candidate generation does not eliminate too many correct candidates.
Table~\ref{tab:performance_comparison} shows the performance of our model and the baselines.
Besides accuracy, we also compute a binomial confidence interval for each model (at a confidence level of 0.02), based on the total number of mentions and the number of correctly mapped mentions. 
The best results are shown in bold text, and all performances that are within the error margin of the best-performing model are shown in gray.
We first observe that, for each dataset, several methods perform within the margin of the best-performing model. However, only two models are consistently within the margin across all datasets: BERT and our method.
Adding extra features (prior, context, coherence) to our base model yields a small increase on the three datasets. 
However, overall, even our base model achieves a performance that is statistically indistinguishable from the state of the art.

\begin{table}[!t]  
	\centering  
	%\fontsize{6.5}{8}\selectfont 
	\small
	\begin{threeparttable} 
		\begin{tabular}{cccc}  
			\toprule  
			\multirow{1}{*}{Model}&ShARe/CLEF&NCBI&ADR\cr    
			\midrule
			- Character Feature  &-1.21&-0.31&-0.30\cr %&-1.21&88.76&92.59\cr
			- Alignment Layer & \underline{-3.80}&\underline{-4.06}&\underline{-3.17}\cr  %&86.30&85.01&89.72\cr
			- CNN Layer &-1.87&-0.93&-0.35\cr %&88.23&88.14&92.54\cr
			\rowcolor{lightgray!50} Our Base Method &90.10&89.07&92.63\cr
			+ Mention-Entity Prior &+0.33&+0.04&+0.03\cr
			+ Context &-0.09&+0.21&-0.24\cr
			+ Coherence  &-0.02&+0.27&+0.11\cr
			\bottomrule  
		\end{tabular}
		\caption{Ablation study}\label{tab:ablation}
	\end{threeparttable}  
\end{table}

\begin{table*}[!t]  
    \centering 
	%\small%
	%\begin{minipage}{.25\linewidth}%
	%	\caption{Performance in the face of typos: Simulated ADR Datasets}\label{tab:simulate}
	%\end{minipage}%
	%\hfill%
	%\begin{minipage}{.72\linewidth}%
	\begin{threeparttable} 
		\begin{tabular}{ccccccc}  
			\toprule  
			\multirow{1}{*}{Model}&Original ADR&10\%&30\%&50\%&70\%&90\%\cr    
			\midrule
			+ Ordering Change  &92.63&92.20&92.18&91.95&92.31&92.05\cr 
			+ Typo &92.63&92.03&91.61&91.38&91.41&91.13\cr  
			\bottomrule  
		\end{tabular}
	\caption{Performance in the face of typos: Simulated ADR Datasets}\label{tab:simulate}
	\end{threeparttable}  
	%\end{minipage}%
\end{table*}

\ignore{
\begin{table*}[!t]  
	%\fontsize{6.5}{8}\selectfont
	%\small
	%\begin{minipage}{.25\linewidth}%
	%			\caption{Model parameter numbers and external resources used.}
	%	\label{tab:parameters}
	%\end{minipage}%
	%\hfill%
	%\begin{minipage}{.72\linewidth}%
	\begin{threeparttable} 
		\begin{tabular}{cccc}  
			\toprule  
			&Sieve-based Model&BERT-based Ranking&Our Base Model\cr    
			\midrule
			Parameter Numbers &-&110M/340M&6.5M/4.9M/2.3M\cr
			Abbreviation Expansion Tool &$\checkmark$&$\checkmark$&$\checkmark$\cr
			Abbreviation Dictionary &$\checkmark$&$\checkmark$&$\checkmark$\cr
			Numeral Dictionary &$\checkmark$&$\checkmark$&$\checkmark$\cr
			Synonym Dictionary&$\checkmark$&$\times$&$\times$\cr
			Spelling Check Dictionary&$\times$&$\checkmark$&$\times$\cr
			Stemming Tool&$\checkmark$&$\checkmark$&$\times$\cr
			Information Retrieval Tool &$\times$&$\checkmark$&$\times$\cr
			\bottomrule  
		\end{tabular}
	\caption{Model parameter numbers and external resources used.}
	\end{threeparttable}
	%\end{minipage}%
\end{table*}
}

\begin{table*}[!t] 
    \centering 
	%\fontsize{6.5}{8}\selectfont
	\small
	%\begin{minipage}{.18\linewidth}%
	%	\caption{Number of model parameters and observed prediction time.}
	%\end{minipage}%
	%\hfill%
	%\begin{minipage}{.81\linewidth}%
	\begin{threeparttable} 
		\begin{tabular}{cccccccccc}  
			\toprule  
			Model&Parameters&\multicolumn{2}{c}{ShARe/CLEF}&\multicolumn{2}{c}{NCBI}&\multicolumn{2}{c}{ADR}&Avg&Speedup\cr
			&&CPU &GPU  &CPU  &GPU  &CPU  &GPU && \cr  
			\midrule
			BERT (large)&340M&2230s&1551s&353s&285s&2736s&1968s&1521s&12.3x\cr
			BERT (base)&110M&1847s&446s&443s&83s&1666s&605s&848s&6.4x\cr
			TinyBERT$_{6}$&67M&1618s&255s&344s&42s&2192s&322s&796s&6.0x\cr
			MobileBERT (base)&25.3M&1202s&330s&322s&58s&1562s&419s&649s&4.7x\cr
			ALBERT (base)&12M&836s&\textbf{129s}&101s&24s&1192s&170s&409s&2.6x\cr    
			Our Base Model&4.6M&\textbf{181s}&131s&\textbf{38s}&\textbf{22s}&\textbf{196s}&\textbf{116s}&\textbf{114s}&-\cr
			\bottomrule  
		\end{tabular}
	\caption{Number of model parameters and observed inference time}	\label{tab:running time}
	\end{threeparttable} 
	%\end{minipage}%
\end{table*}

\subsection*{Ablation Study}
To understand the effect of each component of our model, we measured the performance of our model when individual components are removed or added.
The results of this ablation study on all three datasets are shown in Table~\ref{tab:ablation}.
The gray row is the accuracy of our base model. The removal of the components of the base model is shown above the gray line; the addition of extra features (see the section of  \nameref{sec:extra}) below.
If we remove the Alignment Layer (underlined), the accuracy drops the most, with up to 4.06 percentage points.
This indicates that the alignment layer can effectively capture the similarity of the corresponding parts of mentions and entity names.
The CNN Layer extracts the key components of the names, and removing this part causes a drop of up to 1.87 percentage points. 
The character-level feature captures morphological variations, and removing it results in a decrease of up to 1.21 percentage points. 
%We see that removing any of these components would move the performance of our model out of the error margin of the best-performing model in Table~\ref{tab:performance_comparison} on at least one dataset.
Therefore, we conclude that all components of our base model are necessary.

Let us now turn to the effect of the extra features of our model.
The Mention-Entity Prior can bring a small improvement, because it helps with ambiguous mentions, which occupy only a small portion of the dataset.
The context feature, likewise, can achieve a small increase on the NCBI dataset. On the other datasets, however, the feature has a negative impact. We believe that this is because the documents in the NCBI datasets are PubMed abstracts, which have more relevant and informative contexts. 
The documents in the ShARe/CLEF and ADR datasets, in contrast, are more like semi-structured text with a lot of tabular data. Thus, the context around a mention in these documents is less helpful.
The coherence feature brings only slight improvements. This could be because our method of estimating co-occurrence is rather coarse-grained, 
and the naive string matching we use may generate errors and omissions.
%On the other hand, biomedical entity candidates cannot be distinguished well by context and coherence, because they generally belong to the same sub-class.
In conclusion, the extra features do bring a small improvement, and they are thus an interesting direction of future work. However, our simple base model is fully sufficient to achieve state-of-the-art performance already.

\subsection*{Performance in the Face of Typos}
To reveal how our base model works, we further evaluate it on simulated ADR datasets.  We generate two simulated datasets by randomly adding typos and changing word orderings of mention names.
As described in Table~\ref{tab:simulate}, as we gradually add typos, the accuracy does not drop too much, and adding 90\% of typos only results in a 1.5 percent drop.
This shows our model can deal well with morphological variations of biomedical names.
Besides, ordering changes almost have no effect on our base model, which means it can capture correspondences between mention and entity names.

\subsection*{Parameters and Inference Time}
To measure the simplicity of our base model, we analyze two dimensions: the number of model parameters and the practical inference time.
In Table~\ref{tab:running time}, we compare our model with BERT models, including three popular lightweight models: ALBERT\cite{lan2019albert}, TinyBERT\cite{jiao2019tinybert}, and MobileBert\cite{sun2020mobilebert}.
Although ALBERT's size is close to our model, its performance is still 2.2 percentage points lower than the BERT$_{\textit{BASE}}$
model on average.

The second column in the table shows the number of parameters of different models.
Our model uses an average of only 4.6M parameters across the three data sets, which is 1.6x to 72.9x smaller than the other models.
The third column to the tenth column show the practical inference time of the models on the CPU and GPU.
The CPU is described in the \nameref{sec:experimental setting}, and the GPU we used is a single NVIDIA Tesla V100 (32G).
Our model is consistently the fastest across all three datasets, both for CPU and GPU (except in the fourth column).
On average, our model is 6.4x faster than other BERT models, and our model is much lighter on the CPU.
%has a much bigger advantage on the CPU.

\subsection*{Model Performance as Data Grows}

\begin{figure}[t]
	\centering
	\includegraphics[width=0.45\textwidth]{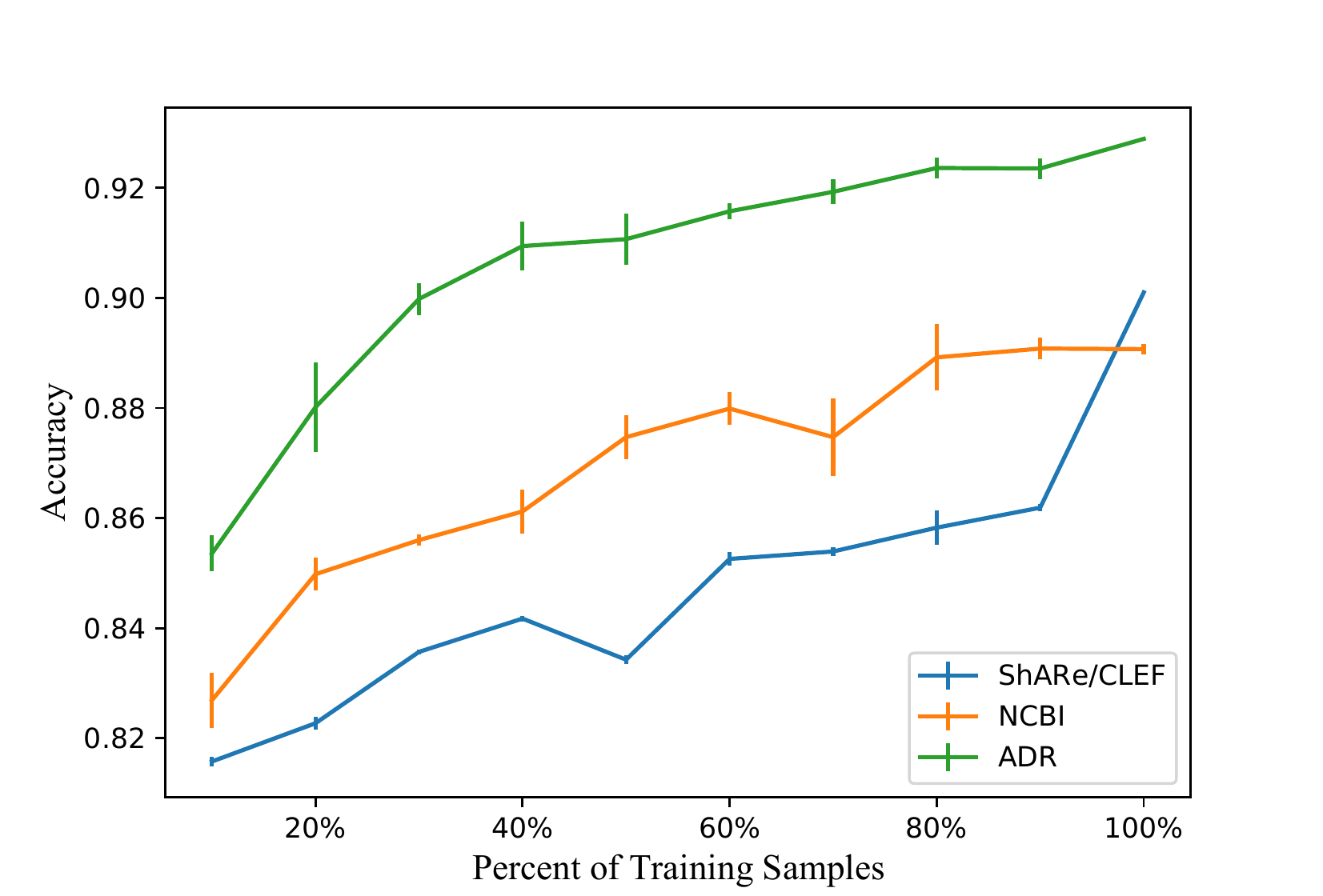}
	\caption{Model efficiency on a small amount of data.}
	\label{fig:smalldata}
\end{figure}

In this section, we study how our model performs with an increasing amount of training samples, by subsampling the datasets.
As shown in Figure~\ref{fig:smalldata}, the performance of our base model keeps growing when we gradually increase the number of training samples. When using 50\% of the training samples, the accuracies of ShARe/CLEF, NCBI, and ADR dataset are already $0.8342, 0.8747,$ and $0.9106$, respectively.
More data leads to better performance, and thus our model is not limited by its expressivity, even though it is very simple.

\section*{Conclusion}
In this paper, we propose a simple and lightweight neural model for biomedical entity linking.
Our experimental results on three standard evaluation benchmarks show that the model is very effective, and achieves a performance that is statistically indistinguishable from the state of the art. BERT-based models, e.g., have 23 times more parameters and require 6.4 times more computing time for inference. Future work to improve the architecture can explore \emph{1)} automatically assigning a weight for each word in the mentions and entity names to capture the importance of each word, depending, e.g., on its grammatical role; \emph{2)} Graph Convolutional Networks (GCNs) \cite{kipf2016semi,wu2020dynamic} to capture graph structure across mentions and improve our notion of entity coherence.

\goodbreak
\section*{Acknowledgments} 
This project was partially funded by the DirtyData project (ANR-17-CE23-0018-01).

\bibliography{aaai21}

\end{document}